\documentclass{article}

\usepackage{microtype}
\usepackage{graphicx}
\usepackage{booktabs} 

\usepackage{multirow}
\usepackage{colortbl} \usepackage{xcolor}   \usepackage{pgfplots}
\usepackage{pgfmath}
\usepackage{pgf}
\usepackage{highlight}
\definecolor{modelnames}{HTML}{f7f4eb}
\definecolor{domains}{HTML}{e3e2de}
\definecolor{darker}{HTML}{bdbcb9}
\usepackage{subcaption}
\definecolor{marinadark}{HTML}{dce3fc}
\definecolor{marinalight}{HTML}{f7f0f7}
\definecolor{validly}{HTML}{9CCC65} \definecolor{invalidly}{HTML}{E57373} \usepackage{listings}
\usepackage{xcolor}

\usepackage{hyperref}

\usepackage[accepted]{icml2025}

\usepackage{amsmath}
\usepackage{amssymb}
\usepackage{mathtools}
\usepackage{amsthm}

\usepackage[capitalize,noabbrev]{cleveref}

\theoremstyle{plain}
\newtheorem{theorem}{Theorem}[section]

\theoremstyle{definition}
\newtheorem{definition}[theorem]{Definition}

\theoremstyle{remark}

\DeclareRobustCommand{\parhead}[1]{\textbf{#1}~}

\usepackage[textsize=tiny]{todonotes}

\icmltitlerunning{Potemkin Understanding in Large Language Models}

\begin{document}

\twocolumn[
\icmltitle{Potemkin Understanding in Large Language Models}

\icmlsetsymbol{equal}{*}

\begin{icmlauthorlist}
\icmlauthor{Marina Mancoridis}{mit}
\icmlauthor{Keyon Vafa}{keyon}
\icmlauthor{Bec Weeks}{bec}
\icmlauthor{Sendhil Mullainathan}{mit}
\end{icmlauthorlist}

\icmlaffiliation{mit}{Massachusetts Institute of Technology}
\icmlaffiliation{keyon}{Harvard University}
\icmlaffiliation{bec}{University of Chicago}

\icmlcorrespondingauthor{Keyon Vafa}{kvafa@g.harvard.edu}
\icmlcorrespondingauthor{Marina Mancoridis}{marinam@mit.edu}

\icmlkeywords{large language models, benchmarks, concept understanding}

\vskip 0.3in
]

\printAffiliationsAndNotice{}

\begin{abstract}
Large language models (LLMs) are regularly evaluated using benchmark datasets. But what justifies making inferences about an LLM's capabilities based on its answers to a curated set of questions? This paper first introduces a formal framework to address this question. The key is to note that the benchmarks used to test LLMs---such as AP exams---are also those used to test people. However, this raises an implication: these benchmarks are only valid tests if LLMs misunderstand concepts in ways that mirror human misunderstandings. Otherwise, success on benchmarks only demonstrates \textbf{potemkin understanding:} the illusion of understanding driven by answers irreconcilable with how any human would interpret a concept. We present two procedures for quantifying the existence of potemkins: one using a specially designed benchmark in three domains, the other using a general procedure that provides a lower-bound on their prevalence. We find that potemkins are ubiquitous across models, tasks, and domains. We also find that these failures reflect not just incorrect understanding, but deeper internal incoherence in concept representations.

\end{abstract}

 \section{Introduction}
\label{sec:intro}

There has been a change in how machine learning evaluations are interpreted. Today, large language models (LLMs) are evaluated on benchmark datasets: curated questions with rubrics for grading. Success on these questions is interpreted as evidence of broader conceptual understanding, which has led to the creation of domain-specific benchmarks across a wide range of concepts \citep{bubeck2023sparks,singhal2023large,guo2023can,liu2024mathbench}. In contrast, while benchmarks were also used to evaluate supervised learning models, success was interpreted differently. For example, if a pathology classifier performs well on X-ray classification, it is not credited with an understanding of vision---we only draw inferences for its performance on specific distributions, inferences that are limited by distribution shift.

In this paper, we first present a framework that clarifies which assumptions are needed to justify interpreting benchmarks in this way for LLMs. 
The framework relies on a key observation: 
these same benchmarks --- AP exams, AIME math competitions, coding challenges --- are widely trusted as tests of understanding in \textit{people}. 
Our framework formalizes why benchmarks reliably measure understanding in people: 
the set of ways in which a human might misunderstand a concept is small and structured.
Benchmark questions are designed to exploit this fact: 
no one can pass a well-designed game theory exam without understanding Nash equilibria or dominated strategies.

\begin{figure}[t]
    \centering
\includegraphics[width=0.37\textwidth]{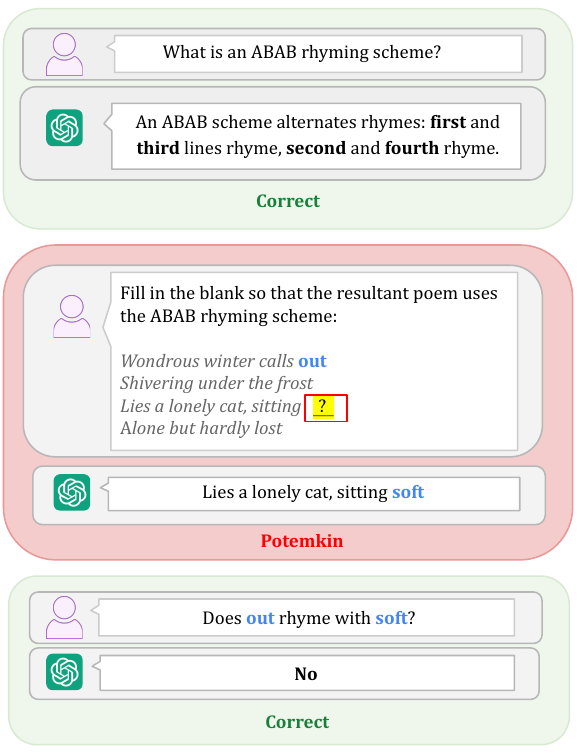}
    \caption{\emph{Illustration of potemkin understanding in a large language model}. This example shows GPT-4o's failure to apply its own conceptual explanation of an ABAB rhyming scheme.}
    \label{fig:mem_examples} 
\end{figure}

This framework also raises an implication: 
that benchmarks designed for humans 
are only valid tests for LLMs if the space of LLM misunderstandings is structured in the same way as the space of human misunderstandings. If LLM misunderstandings diverge from human patterns, models can succeed on benchmarks without understanding the underlying concepts. When this happens, it results in pathologies we call \textbf{potemkins}\footnote{This term comes from Potemkin villages---elaborate facades built to create the illusion of substance where none actually exists.}.

\Cref{fig:mem_examples} illustrates a potemkin. When an LLM is asked to explain an ABAB rhyming scheme, its response is clear and correct (top panel). At first glance, it may appear that the LLM has understood the concept, in the same way that a human with the provided explanation would understand. However, when tasked to generate text in an ABAB rhyming scheme, the LLM fails, producing non-rhyming words (middle panel). Moreover, the LLM seems to recognize that its output does not rhyme (bottom panel). 
This specific combination of correct and incorrect answers is irreconcilable with any answer that a human would give.

Potemkins occur when an LLM performs well on tasks that would indicate conceptual understanding if a human completed them, but do not indicate understanding in the LLM. This paper develops two procedures for measuring the prevalence of potemkins in LLMs. The first is tailored to a specific kind of potemkin: the divide between an LLM's ability to explain a concept and apply it. We collect a benchmark dataset across three domains --- literary techniques, game theory, and psychological biases --- designed to measure the prevalence of this type of potemkins. In contrast, the second procedure is general and doesn't make assumptions on the structure or domain of potemkins, but it only provides a lower-bound on their prevalence. 

We apply these procedures to a set of LLMs and find that potemkins are ubiquitous. For example, despite models being able to define concepts in each domain in our benchmark dataset near-perfectly, they struggle to apply these concepts accurately. We find that potemkins are not arising due merely to incorrect understanding of concepts, but rather due to incoherence.
Despite the fact that the automated procedure provides only a lower bound, it still identifies high rates of potemkins across LLMs.

The rest of the paper is structured as follows: \Cref{sec:framework} presents our framework for defining potemkins. \Cref{sec:benchmark} introduces a benchmark designed to measure the presence of potemkins and shows their ubiquity across key domains. \Cref{sec:automatic_eval} corroborates these findings using a separate, automated evaluation procedure.

\section{Framework}
\label{sec:framework}
Potemkins arise when there is a misalignment between how humans and large language models (LLMs) understand concepts. Here, we present a theoretical framework for defining conceptual understanding. 

\parhead{Conceptual understanding in people.}
At a high level, a concept consists of a set of rules that describe objects. For example, a haiku is a concept, consisting of logic that can be used to classify poems. Meanwhile a fact like ``Abraham Lincoln was born in 1809'' is not a concept because it does not correspond to any set of generalizable rules.

What does it mean for a human to understand a concept? 
The full scope of possible ways to demonstrate understanding is enormous; for example, we may expect someone who understands the concept of a haiku to be able to define it, generate haikus about arbitrary topics, and categorize every poem as a haiku or otherwise. Asking people to enumerate every possible example of a concept is intractable. Instead, it typically suffices to demonstrate understanding with a few specific examples. For example, if someone can accurately define a haiku, we would have confidence that they have understood the concept of a haiku.

Why is it reasonable to infer that people have understood a concept after only seeing a few examples? The key insight is that while there exist a theoretically very large number of ways in which humans might misunderstand a concept, only a limited number of these misunderstandings occur in practice. This is because people misunderstand concepts in structured ways. For example, if a person mistakenly believes that haikus follow a 5-8-5 syllabic structure, the examples of haikus they come up with will all be incorrect in the same way. This logic underlies why we use exams to test conceptual understanding in people: even though SAT and AP exams only consist of a tiny fraction of possible questions about a concept, the questions are structured so that conceptual understanding is necessary to achieve high scores. The space of human misunderstandings is predictable and sparse.

We formalize this notion by defining $\mathcal X$ as the set of all strings that are relevant to a concept: for example, a string can correspond to a possible definition of a concept or a possible example of one. Not every string that's relevant to a concept is a valid use of it. An \textbf{interpretation} of a concept is defined to be any function $f: \mathcal X \to \{0, 1\}$, where the output indicates whether the string is considered valid in the interpretation (0 for invalid, 1 for valid). There is a single correct interpretation, denoted by $f^*$. The set of possible ways for humans to interpret concepts is denoted by $\mathcal F_h$. Every function $f \in \mathcal F_h$ that is not equal to $f^*$ corresponds to a way in which a human might misunderstand a concept. 

\begin{figure*}
    \centering
    \includegraphics[width=1.0\textwidth]{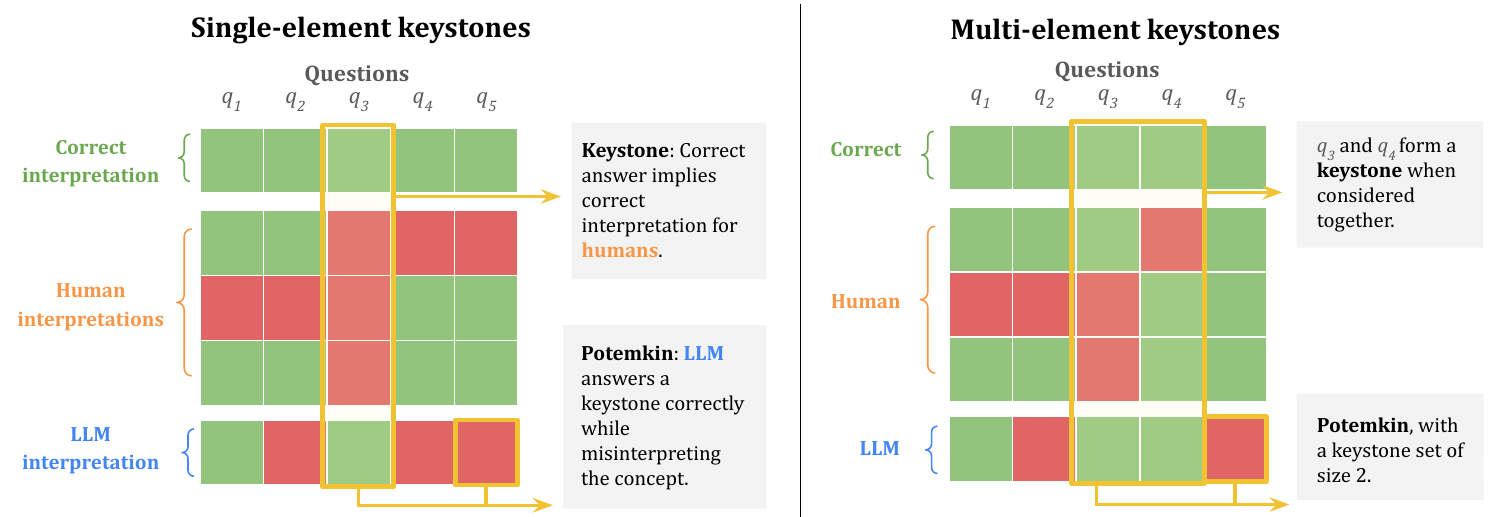}
    \caption{A schematic representation of keystones and potemkins. Rows represent interpretations of a concept and columns represent questions. Questions can either be interpreted \textbf{\textcolor{validly}{correctly}} or \textbf{\textcolor{invalidly}{incorrectly}}. A keystone is a set of questions that can only be interpreted correctly by a human who has understood the concept. An LLM has potemkin understanding when it correctly interprets all questions in a keystone but does not understand a concept. 
    }
    \label{fig:keystone_set} 
\end{figure*}

Consider one possible way $f \in \mathcal F_h$ that a human might interpret a concept. 
How would we check whether $f$ is the correct interpretation? 
In practice, it is intractable to check whether $f(x) = f^*(x)$ across all strings $x \in \mathcal X$. 
Instead, we would like to check whether $f(x) = f^*(x)$ on a few strings $x$. 
But when is this activity justified? The answer is revealed by the framework: we can test conceptual understanding in people using a curated set of examples whenever those examples are chosen so that they are only interpreted correctly by people who have understood the concept.

Formally, we define a \textbf{keystone} set $\mathcal S \subseteq \mathcal X$ as a minimal set of instances such that if $f \in \mathcal F_h$ and $f(x) = f^*(x)$ for all $x \in \mathcal S$, then $f = f^*$. That is, if every example in the keystone set is interpreted correctly, it cannot be reconciled with any misconceived human notion of the concept. See \Cref{fig:keystone_set} for a visual depiction of a keystone set.

This approach shows why testing conceptual understanding in people is feasible.  Testing understanding of a concept doesn't require testing people for all relevant examples. Instead, it only requires testing examples in a keystone set.

\parhead{Potemkins invalidate LLM benchmarks.}
Humans are evaluated on tests that are designed to highlight keystones --- questions that can only be answered correctly by a person who has fully understood a concept. 
Common benchmarks for evaluating LLMs include questions sourced from AP exams, SAT tests, medical examinations, and other standardized assessments \cite{hendrycks2020measuring, clark2016my, jin2021disease}.

But how effective are tests that are designed for people when it comes to evaluating LLMs?
To answer this question, define $\mathcal F_l$ as the set of ways for any possible LLM to interpret a concept, where each $f \in \mathcal F_l$ is an interpretation $f: \mathcal X \to \{0, 1\}$.\footnote{We don't impose a single way to translate an LLM's output into a statement about correctness; for example, an LLM might be instructed to return 0 if a particular instance is incorrect and 1 if it's correct.}

\begin{definition}
An LLM has \textbf{potemkin understanding} if its interpretation satisfies $f(x) = f^*(x)$ for all $x$ in a keystone $S$, but $f \neq f^*$. In this case, we refer to any $x$ such that $f(x) \neq f^*(x)$ as a \textbf{potemkin}.
\end{definition}

In other words, potemkins arise when an LLM answers keystone questions correctly but does not have the correct interpretation of a concept. The prevalence of potemkins in LLMs is consequential for benchmarking:

\parhead{Result:} Keystones are a valid way to test conceptual understanding in LLMs if $\mathcal F_l = \mathcal F_h$. 

\parhead{Corollary:} If an LLM has potemkin understanding, it illustrates that keystones are invalid tests of LLM understanding.

These results imply that benchmarks based on keystone questions for people are invalid tests for LLMs if LLMs are capable of potemkin understanding. Suppose we were to test an LLM like we would a human: by testing whether it correctly answers the questions in a keystone set. For example, we may prompt it with the questions on an AP exam and measure whether it answers them correctly. A human that performs well on these questions must have the correct interpretation $f^*$ because their interpretations are limited to $\mathcal F_h$. However, an LLM that has not understood the concept may still perform well on the exam if $\mathcal F_l \neq \mathcal F_h$; that is, if it is capable of misinterpreting a concept in ways that do not mirror human misinterpretations.

We note that under the hold-out principle, performance on benchmark questions still guarantees performance on other questions drawn from the same distribution; if an LLM scores 95\% accuracy on an i.i.d. sample of AP questions, it is expected to score 95\% on a separate sample. 
However, analogous to how the hold-out principle only holds under i.i.d. data assumptions, success on keystone questions only guarantees success on other relevant tasks under a ``no potemkins'' assumption. It is on these other tasks --- not standardized tests --- that we typically care about LLM performance in the real world. 

Thus, to evaluate whether LLM benchmarks work, we must first determine the prevalence of potemkins. The remainder of the paper is dedicated to developing procedures to quantify the prevalence of potemkins.

 \section{A Benchmark Dataset for Potemkins}
\label{sec:benchmark}

This paper presents two procedures for measuring the prevalence of potemkins in large language models (LLMs). 
This section describes one procedure, based on a benchmark dataset we collect that measures a specific kind of potemkin failure: a divide between describing and applying concepts. Specifically, we construct a dataset spanning $32$ concepts from $3$ distinct domains: literary techniques, game theory, and psychological biases, collecting $3,159$ labeled data points. We find that even when models can correctly define a concept, they often fail to accurately apply it in classification, generation, and editing tasks. All collected data, annotations, and analysis are made publicly available at the Potemkin Benchmark Repository.\footnote{\url{https://github.com/MarinaMancoridis/PotemkinBenchmark.git}}

\subsection{Benchmark motivation}
\label{sub:benchmark_design}

We design a benchmark that measures a specific kind of potemkin. We first note that any human who correctly answers the keystone questions for a concept must be able to correctly \emph{use} that concept in a concrete instance. This is because, by definition, keystone success in humans indicates correct concept understanding. 
Thus, we identify potemkins in LLMs as instances where (1) the LLM can correctly answer keystones but (2) it fails to correctly use that same concept in a concrete instance.

What might we use as our keystone? A common keystone for a concept is  \textbf{definitions}; we have faith that humans who can clearly define the concept of a haiku have understood haikus. Thus, a potemkin occurs when an LLM that can define a concept correctly cannot use it. 

What would it mean for an LLM to be able to use a concept in a concrete instance? We consider three such tasks, each that offers a unique perspective for measuring potemkins. One task is \textbf{classification}: answering whether an example is a correct application of a concept. Another task is \textbf{generation}: producing an instance of a concept that adheres to specific constraints. The last task we consider is \textbf{editing}: modifying an example so that it either belongs or doesn't belong to a concept. We provide more details for these tasks in \Cref{sub:tasks}; see  Appendix Section \ref{app:framework_specs} for a visual representation of our experiment.

What it means to use a concept depends on the domain being tested. We choose concepts from a diverse array of domains: \textbf{literary techniques}, \textbf{game theory}, and \textbf{psychological biases}. These domains together span generative linguistics, formal constructs, and human understanding. We examine concepts like ``\emph{analogy}" in literary techniques, ``\emph{Pareto optimality}" in game theory, and ``\emph{sunk cost fallacy}" in psychological biases. We explore a total of $32$ distinct concepts within the domains, with a full list provided in Appendix Section \ref{appendix:concept_choices}. Given the diversity of our concepts and tasks, evidence of potemkins in our analysis would suggest not an isolated issue but a systemic category of failure in LLMs.

\subsection{Benchmark construction.}
\label{sub:tasks}
We construct datasets to evaluate concept explanation and concept use in each of the three domains. 
To achieve this, we construct datasets for each of the three domains, varying generation and evaluation methods to enhance the robustness of our findings. 
Collecting data or validating responses from some domains requires hand-labeling, while others are automatic. To ensure high-quality annotations, we rely on a mix of domain experts and paper authors to evaluate model responses. Our analysis spans the following $7$ models: Llama-3.3 (70B), GPT-4o, Gemini-2.0 (Flash), Claude-3.5 (Sonnet), DeepSeek-V3, DeepSeek-R1, and Qwen2-VL (72B). Model names are abbreviated in subsequent tables. Below, we describe our data collection process for each of the four tasks.

\parhead{Definition.} 
To assess whether LLMs can explain concepts, we prompted models to define a concept in a given domain. We prompted each of $7$ models to define each of $32$ concepts, resulting in a total of $224$ generated definitions across domains. We evaluated the definitions ourselves, as some of the concepts required specialized knowledge to evaluate accurately. For example, evaluating the accuracy of a definition for the literary techniques concept of a ``Shakespearean sonnet'' required confirming that it accurately described the ABAB CDCD EFEF GG rhyming scheme and specified that the poem must be entirely written in iambic pentameter.

\parhead{Classification.} 
In classification tasks, models must determine if presented examples are valid instances of a given concept. The model is given an instance and asked: ``Is the following example a true instance of the concept $c$?" For example, to assess the model’s grasp of the concept of ``slant rhyme'', we could present pairs of words---some that rhyme and others that do not---and ask the model to evaluate whether they qualify.

Evaluating a model’s ability to classify concepts requires creating positive and negative examples. We vary data generation approaches across domains. For \emph{literary techniques}, we crafted original examples and collected online examples from recent poetry competitions (ensuring they post-date each LLM’s training cutoff). As the \emph{game theory} domain requires specialized knowledge, we recruited Economics PhD students to produce true and false instances. For the \emph{psychological biases} domain, we gathered $40$ text responses from Reddit’s \emph{“r/AmIOverreacting”} thread, annotated by expert behavioral scientists recruited via Upwork. Models classified each example, and we compared their outputs against the label. Overall, we generated 2,030 annotations. For further details, see Appendix Sections \ref{sec:app_specifications_classification} and \ref{sec:app_expert_psych_survey}.

\parhead{Constrained generation.} This task assesses a model’s ability to use concepts by requiring it to generate examples adhering to specific constraints. This tests the model’s capacity to flexibly apply concepts within structured boundaries. For instance, for the concept of ``strict dominance" in game theory, we might ask the model to construct an example specifying constraints like the number of players or which players have strictly dominant strategies. We defined different constraints for each concept and varied annotation techniques across domains: 
literature labels came from paper authors, game theory labels were generated automatically, and psychological biases labels were determined by majority expert consensus recruited via Upwork. Overall, 224 model responses were evaluated. See Appendix Sections \ref{sec:app_specifications_generation} and \ref{sec_app:psych_generate_survey} for detailed specifications.

\parhead{Editing.} This task evaluates a model’s ability to use concepts by requiring it to identify modifications that could transform an instance into either a true or false example of a given concept. For instance, we might present the model with a partially obscured haiku—where a section of the poem is missing—and ask what could be added to complete the poem and ensure it qualifies as a true instance of a haiku. To assess performance, we prompt the model with an instance to edit and evaluate its response. Our prompting and evaluation strategies were tailored to each domain. In total, we gathered 791 annotations of model responses. For further specifications, see in Appendix Section \ref{sec:app_specifications_editing}.

\subsection{Results}
\label{sub:results}
We analyze $7$ large language models across $32$ concepts. 
These models were chosen for their popularity and range of developers and sizes. We collect inferences using APIs from OpenAI, Together.AI, Anthropic, and Google.

\begin{table}[t]
\centering
\begin{tabular}{lccc}
\toprule
& \multicolumn{3}{c}{Potemkin rate, as measured by:} \\
\cmidrule(lr){2-4}
\textbf{Model} & \textbf{Classify} & \textbf{Generate} & \textbf{Edit} \\ 
\midrule
Llama-3.3           & 0.57 (0.06) & 0.43 (0.09) & 0.36 (0.05) \\
Claude-3.5          & 0.49 (0.05) & 0.23 (0.08) & 0.29 (0.04) \\
GPT-4o              & 0.53 (0.05) & 0.38 (0.09) & 0.35 (0.05) \\
Gemini-2.0          & 0.54 (0.05) & 0.41 (0.09) & 0.43 (0.05) \\
DeepSeek-V3         & 0.57 (0.05) & 0.38 (0.09) & 0.36 (0.05) \\
DeepSeek-R1         & 0.47 (0.05) & 0.39 (0.09) & 0.52 (0.05) \\
Qwen2-VL            & 0.66 (0.06) & 0.62 (0.09) & 0.52 (0.05) \\
\midrule
\textbf{Overall}    & \textbf{0.55 (0.02)} & \textbf{0.40 (0.03)} & \textbf{0.40 (0.02)} \\
\bottomrule
\end{tabular}
\caption{\emph{Potemkin rate on classify, generate, and edit tasks}.
The potemkin rate is the percent of questions about a concept that are answered incorrectly, conditional on a model defining that concept correctly.  We scale the classification values so that chance-level performance corresponds with a potemkin rate of $1$. Standard errors are in parentheses.}
\label{tab:main_results_models_only}
\end{table}

For each (model, concept) pair, we first determine whether the model provides a correct definition. If so, we evaluate its accuracy on the three additional tasks: classification, generation, and editing. Responses are labeled as correct or incorrect according to our framework specifications.

We measure the \textbf{potemkin rate} exhibited by models. We define the potemkin rate of a model as the proportion of questions that the model solves incorrectly when it solves a keystone correctly. 
For tasks with a chance accuracy of $0.50$, we scale this value by a factor of $2$, so that a potemkin rate of $1$ corresponds to chance-level performance.

Our findings reveal high potemkin rates across all models and domains, as summarized in \Cref{tab:main_results_models_only}. Models define concepts correctly 94.2\% of the time. However, conditioned on correct definitions, their performance sharply decreases when tasked with \emph{using} those concepts, as exhibited by the high potemkin rates in the table.

\begin{figure*}[h] \centering
    \includegraphics[width=1.0\textwidth]{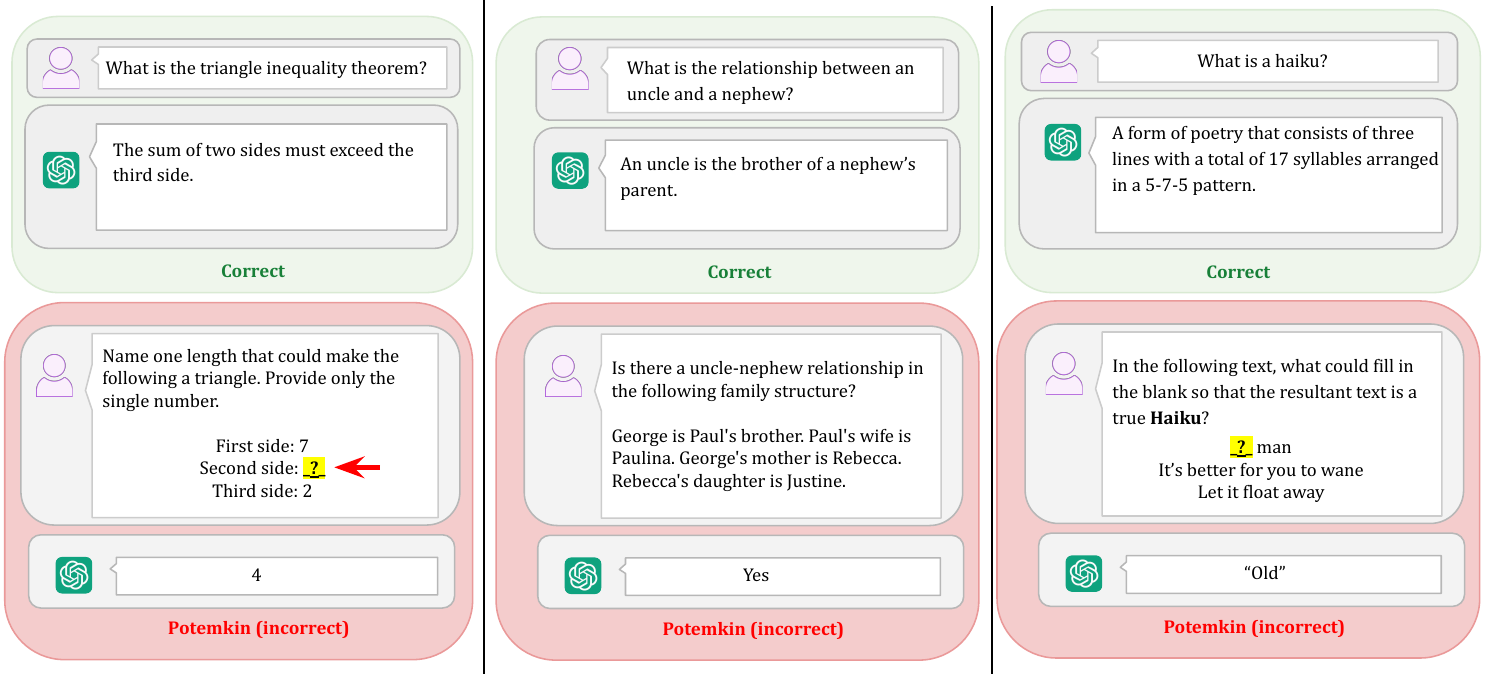}
    \caption{\emph{Examples of potemkins.} In each example, GPT-4o correctly explains a concept but fails to correctly use it.}
    \label{fig:more_potemkin_examples} \end{figure*}

While performance varies slightly across models and tasks, 
we find that \emph{\textbf{potemkins are ubiquitous}} across all models, concepts, and domains that we analyzed. See Table \ref{tab:main_results_2} (Appendix Section \ref{sec:app_full_benchmark_results}) for further details of our results. Further examples of potemkins are provided in Figure \ref{fig:more_potemkin_examples}. 

\parhead{Discussion.} 
We raise two points of discussion related to our benchmark. 
One possible concern is that while our benchmark only uses single definition questions as part of the keystone set, in reality keystones may include multiple questions, not only about defining the concept. For example, we may not have faith that a student who defines the quadratic formula truly understands what it is until we see them apply it a few times. 

\begin{figure} 
    \centering
    \includegraphics[width=0.5\textwidth]{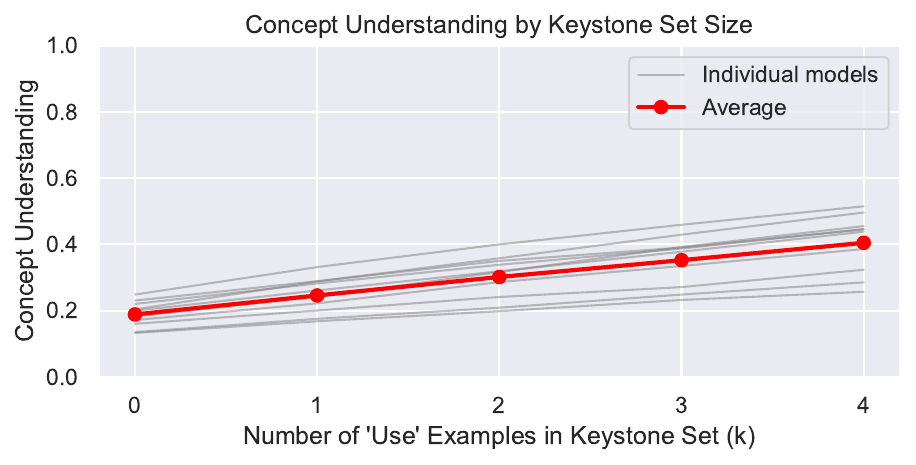}
    \caption{\emph{Impact of expanding keystone sets on concept understanding values.} We extend the keystone from a single definition to include multiple classification (``use'') examples. 
``Concept understanding'' measures whether a model answers 10 additional ``use'' questions after answering keystones correctly, averaged across concepts. 
}
\label{fig:keystone_set_conditioned}
\end{figure}

To address this, we conduct a supplementary analysis that simulates performance when keystones contain additional questions. 
Specifically, for each concept, we consider keystone sets that require not only the correct definition but also $k$ correct responses to classification (``use'') questions. We perform a simulation exercises that measures, among concepts for which an LLM answers the expanded set of keystones correctly, how well it performs on other ``use'' questions. Specifically, we say a model has ``understood a concept'' if it answers 10 additional ``use'' questions after answering keystones correctly. \Cref{fig:keystone_set_conditioned} shows how concept understanding varies with the size of the keystone set (only including models that outperform chance). Expanding the keystone set yields only modest performance gains.

Another concern is whether the \emph{use} tasks we consider in the benchmark are too hard such that humans would fail these as well. Qualitatively, we find this is not the case; Appendix \ref{sec:app_qualitative_potemkins} shows examples of failures that humans who understand the concept would not make.

 \section{Automatically Evaluating Potemkins}
\label{sec:automatic_eval}
In this section we present a different, automated procedure for evaluating the presence of potemkins. 

\subsection{Warmup: Incoherence}
\label{sub:coherence}
\Cref{sec:benchmark} demonstrates that potemkin understanding is ubiquitous in LLMs. 
There are two possibilities for why this might be the case. One possibility is that the LLMs have slightly misaligned but internally consistent understanding of concepts. Another possibility is that their conceptual grasp is incoherent, with conflicting notions of the same idea.

\begin{figure}\centering
    \includegraphics[width=0.5\textwidth]{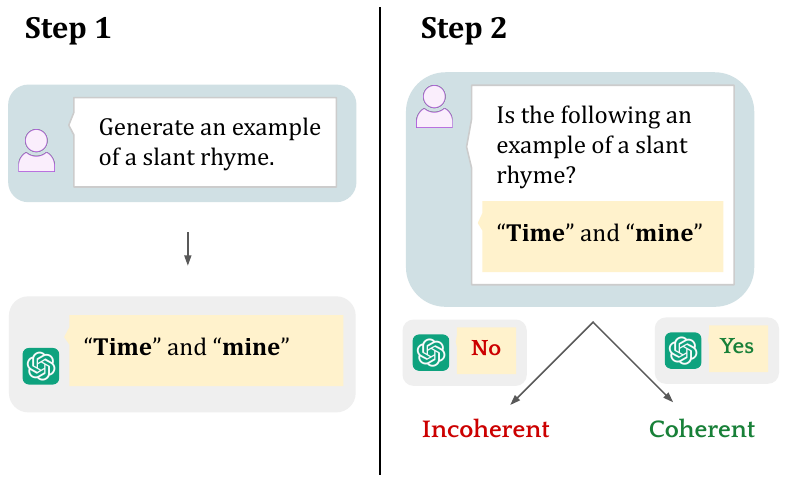}
    \caption{\emph{Illustration of the method for evaluating incoherence in models}. In the first step, the model generates an instance or non-instance of a given concept. In the second step, the model evaluates whether the instance it generated is a true or false example of the concept.}
    \label{fig:coherency_steps} \end{figure}

To distinguish between these two cases, we test specifically for conceptual  incoherence within models. 
We measure incoherence in two steps. First, we prompt a model to generate either an instance or a non-instance of a specific concept (e.g., producing an example of a slant rhyme). Then, we present the model’s generated output back to it (in a separate query), asking whether this output is indeed an instance of the concept. In the slant rhyme example, this means testing if the model recognizes its own example as a slant rhyme. Figure \ref{fig:coherency_steps} summarizes this procedure.

We quantify incoherence by calculating the percentage of cases where the model’s initial generation does not match its subsequent classification. 
We use the same concepts from \Cref{sec:benchmark}. 
Since random-chance accuracy is $0.50$, we then multiply this value $2$, rescaling scores such that $0$ indicates no incoherence and $1$ indicates as-good-as-random performance. 
See Appendix Section \ref{sec:app_incoherence_data_collection} for more details on our data collection process. 

The results are presented in the first column of Table \ref{tab:incoherence_and_auto_eval}. We observe incoherence across all examined models, concepts, and domains, with scores ranging from $0.02$ to $0.64$. Although these scores are better than random, they nonetheless indicate substantial limitations in models' ability to consistently evaluate their own outputs. This indicates that conceptual misunderstandings arise not only from misconceiving concepts, but also from inconsistently using them. A detailed breakdown of incoherence scores by domain is provided in Appendix Section \ref{sec:app_incoherence_by_domain}.

\begin{table}\centering
\begin{tabular}{lcc}
\toprule
\textbf{Model} & \textbf{Incoherence} & \textbf{Potemkin rate} \\
 & & \textit{(lower bound)} \\
\midrule
Llama-3.3     & 0.19 (0.03) & 0.82 (0.02) \\
Claude-3.5    & 0.61 (0.05) & 0.36 (0.02) \\
GPT-4o        & 0.64 (0.05) & 0.46 (0.06) \\
GPT-o1-mini   & 0.16 (0.03) & 0.66 (0.02) \\
GPT-o3-mini   & 0.03 (0.01) & 0.66 (0.04) \\
Gemini-2.0    & 0.09 (0.02) & 0.86 (0.02) \\
DeepSeek-V3   & 0.13 (0.03) & 0.38 (0.02) \\
DeepSeek-R1   & 0.04 (0.02) & 0.50 (0.02) \\
Qwen2-VL      & 0.13 (0.03) & 0.82 (0.00) \\
\midrule
\textbf{Overall} & \textbf{0.22 (0.01)} & \textbf{0.62 (0.01)} \\
\bottomrule
\end{tabular}
\caption{\emph{Incoherence scores and potemkin rates across models.} An incoherence score of $0$ indicates perfect performance and a score of $1$ indicates good-as-random performance. Potemkin rate is defined as $1-$ accuracy, multiplied by $2$ (since random-chance accuracy on this task is $0.5$, implying a baseline potemkin rate of $0.5$). The automatic evaluation procedure provides a lower bound on potemkin rate. Standard errors are in parentheses. }
\label{tab:incoherence_and_auto_eval}
\end{table}

\subsection{A lower bound on potemkin rates.}
\label{sub:automatic_eval_procedure}
Our incoherence analysis used LLM self-grading as a way to measure concept misunderstanding. This insight motivates another, more general automated procedure for measuring potemkins. The idea is as follows: if an LLM understands a concept, its answers to new questions about that concept should be judged as being correct \textit{by the same LLM}. How often the LLM judge disagrees with its original answers reveals the prevalence of potemkins. 

\begin{figure*}[h] \centering
    \includegraphics[width=1.0\textwidth]{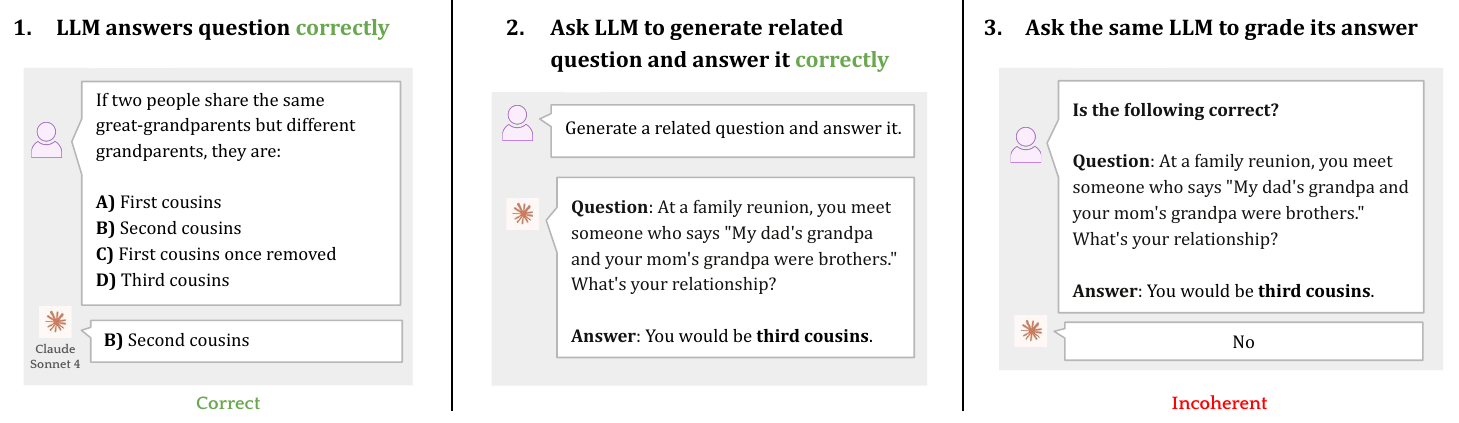}
    \caption{\emph{Example of a potemkin documented by our automatic evaluation procedure.} In this example, the language model fails to agree with its own answer for a question that it generated about the topic of second cousins.}
    \label{fig:automatic_eval_schematic} \end{figure*}

Specifically, we prompt an LLM with questions from a benchmark and automatically grade whether it is correct. If it answers correctly, we prompt it to generate other questions that rely on the same concept. Then, for each question, we prompt the LLM to answer the question correctly and then re-prompt the same LLM to grade its response. Whenever the judge's response deviates from the expected response, it indicates an example of a potemkin. See Figure \ref{fig:automatic_eval_schematic} for an example of this procedure.

Importantly, this approach provides a \textit{lower-bound} on the rate of potemkins. To see why, consider the scenarios in which the LLM judge grades a response as incorrect. If the LLM judge correctly grades an incorrect response as incorrect, that LLM exhibited misunderstanding in responding to the original question. 
If the LLM judge mistakenly grades a correct response as incorrect, it still reveals conceptual misunderstanding in the judge: a judge that understood the concept would've graded it correctly. 
The reason it's a lower bound is that the judge may grade an incorrect response as correct, which would be an example of conceptual misunderstanding but isn't captured by this metric. 
Similarly, generating questions that are unrelated to a concept would also demonstrate misunderstanding, although qualitatively we do not find this to be the case. 

We perform this analysis for the same models as above using questions sampled from MMLU \cite{hendrycks2020measuring}. We prompt LLMs to generate $5$ related questions for each conceptual question it answers correctly and then prompt it to respond to each question. In order to avoid sycophancy in the judge \cite{malmqvist2024sycophancy} (e.g. an LLM may always grade responses as correct), we also prompt the LLM to generate a slightly incorrect response and also re-prompt the LLM to grade it (see Appendix Section \ref{sec:app_auto_eval_prompts} for examples of prompts). Results are reported in Table \ref{tab:incoherence_and_auto_eval}. Our reported potemkin rate is scaled so that $1$ corresponds to chance accuracy and $0$ corresponds to perfect performance. Despite the procedure only providing a lower bound, the potemkin rate is high: $0.62$. These results corroborate the benchmark findings from \Cref{sec:benchmark}.

\section{Related Work}
\parhead{Providing a method to validate benchmarks.} Researchers caution that high scores on standard NLP benchmarks may not be informative of true understanding, highlighting that the “leaderboard mentality” overestimates perceptions of a model’s capabilities \citep{bowman2021will, mitchell2021ai, church2019survey, ribeiro2020beyond, recht2019imagenet, raji2021ai}. Benchmarking raises the challenge of \emph{construct validity}—the degree to which a test measures the concept it claims to measure.

Many challenges undermine the reliability of benchmarks as measures of conceptual understanding. Benchmarks are frequently inadequately developed due to undervalued data collection efforts \cite{gebru2021datasheets, paullada2021data, sambasivan2021everyone}. High scores may instead reflect superficial shortcuts \cite{geirhos2020shortcut, lapuschkin2019unmasking}, while benchmarks often lack sufficient statistical power \cite{card2020little} and aggregate metrics obscure critical failures \cite{wu2019errudite}. These challenges may reflect broader fallacies in measuring intelligence, such as assuming intelligence can be neatly placed on a linear scale or task difficulty uniformly gauged \cite{mitchell2021ai}.

Efforts to improve evaluation approaches include tests of robustness to noise \cite{belinkov2017synthetic, rychalska2019models}, adversarial perturbations \cite{ribeiro2018semantically, iyyer2018adversarial}, logical consistency checks \cite{ribeiro2019red}, focusing on explanations \cite{ribeiro2016should}, and behavioral evaluations through frameworks such as CheckList \cite{ribeiro2020beyond}. There has been a shift from static benchmarks to dynamic, real-world interaction-based assessments, exemplified by ChatBench  \cite{chang2025chatbench}, WildChat \cite{zhao2024wildchat}, ChatbotArena \cite{chiang2024chatbot}, and the human-in-the-loop adversarial framework Dynabench \cite{kiela2021dynabench}. Additionally, best practices in benchmarking have been explored through initiatives like BetterBench \cite{reuel2024betterbench} and comprehensive, multi-metric evaluation methods \cite{liang2022holistic}.

Our paper introduces a framework that clarifies when benchmarks validly measure true concept understanding in LLMs: when there are no potemkins. We provide two procedures for finding potemkins and show potemkins to be ubiquitous in our analysis. The high potemkin rates that we observe invalidate using existing benchmarks as measures of understanding.

\parhead{Providing a framework for evaluating depth of concept understanding.} Researchers have documented numerous \emph{failure modes} in large language models, including limitations in linguistic reasoning \cite{dagan2010recognizing, bowman2015large, dentella2024testing, arkoudas2023gpt, dickinson2003detecting}, common-sense problem solving \cite{shwartz2020neural}, planning \cite{valmeekam2023planning}, alignment with human generalization \cite{vafa2024large}, factual recall and knowledge representation \cite{meng2022locating, geva2020transformer, geva2022transformer, geva2023dissecting, tam2022evaluating, petroni2019language, zhu2020modifying, mitchell2021fast, yao2022kformer}, distractability \cite{shi2023large}, and coherent world-model formation \cite{ivanova2024elements, vafa2024evaluating}. They have identified \emph{logical inconsistencies} through methods including reversed claims \cite{saba2024llms, berglund2023reversal}, decision-making inconsistencies \cite{fluri2024evaluating}, factual correctness in paraphrases \cite{elazar2021measuring}, compositional tasks \cite{press2022measuring}, negation handling \cite{hosseini2021understanding}, common human misconceptions \cite{lin2021truthfulqa}, semantic equivalence \cite{jang2021accurate}, consistency benchmarks \cite{jang2022becel, jang2023consistency}, and validation-generation alignment \cite{li2023benchmarking}.

Our work highlights that models may comprehend concepts in fundamentally different ways than humans. Potemkins reveal modes of conceptual understanding that humans, by construction, cannot exhibit. Our work frames prior methods of evaluating concept understanding as implicit tests for potemkins, and proposes an explicit, systematic approach to uncovering these discrepancies. Potemkins are to conceptual knowledge what hallucinations are to factual knowledge—hallucinations fabricate false facts; potemkins fabricate false conceptual coherence. Yet potemkins pose a greater challenge: hallucinations can be exposed through fact-checking, but potemkins require unraveling subtle inconsistencies in a model’s apparent understanding. By identifying potemkins, we introduce implementable frameworks to assess benchmark validity and conceptual depth, transforming the abstract notion of understanding into actionable tools that can be leveraged for future model improvement.

 \section{Conclusion}
In this paper, we introduce the phenomenon of \emph{potemkin understanding}— a failure mode of large language models (LLMs) whereby apparent comprehension revealed by successful benchmark performance is undermined by non-human patterns of misunderstanding. We began by formalizing a framework for evaluating when benchmarks designed for humans serve as valid tests for understanding in LLMs. This framework reveals that benchmarks are only valid if LLMs misunderstand concepts in the same ways that humans do. If models deviate from human-like misunderstandings, they can still correctly answer keystone questions without truly understanding the underlying concept, producing potemkins.

Through two complementary empirical procedures---one using a new benchmark dataset across literary techniques, game theory, and psychological biases, and the other employing an automated evaluation strategy---we quantified the prevalence of potemkins across a variety of tasks, concepts, domains, and models. Both procedures reveal high potemkin rates, even in models that appear highly capable by conventional benchmark standards. Our tests for incoherence reveal that models contain conflicting representations of the same idea.

Our approach has limitations that suggest avenues for further exploration. The benchmark datasets we developed, while extensive, are not exhaustive. Additional datasets spanning a broader range of concepts and types of keystone questions could enable a more comprehensive identification of potemkins. There is also significant potential in methodologies explicitly designed to detect and reduce potemkin rates in language models. Systematically integrating techniques for potemkin detection and mitigation into existing model training and evaluation pipelines represents a promising direction for future research.

\section*{Acknowledgments}
Marina Mancoridis is supported by the MIT Presidential Graduate Fellowship. Keyon Vafa is supported by the Harvard Data Science Initiative. We also thank Steven Ma, Cassidy Shubatt, and Charlotte Siegmann for helpful comments and feedback.

\section*{Impact Statement}
Our work puts forth formalizations and benchmarks to assess concept understanding in large language models (LLMs). We document \emph{potemkin understanding} as an overlooked but pervasive category of failure in LLMs. By demonstrating that models can correctly answer keystone questions without true concept comprehension, our findings compel a re-examination of the validity of benchmarks widely trusted in AI development. The framework and empirical tools we provide allow researchers and practitioners to systematically identify and address potemkins, paving the way toward the development of deeper and more reliable concept understanding.

This paper presents a dataset using data collected from human surveys. We received an IRB review and exemption for this study.

\bibliography{main}
\bibliographystyle{icml2025}

\newpage
\appendix
\onecolumn
\section{Illustration of the Benchmark Design}
\label{app:framework_specs}
\begin{figure*}[h] \centering
    \includegraphics[width=1.0\textwidth]{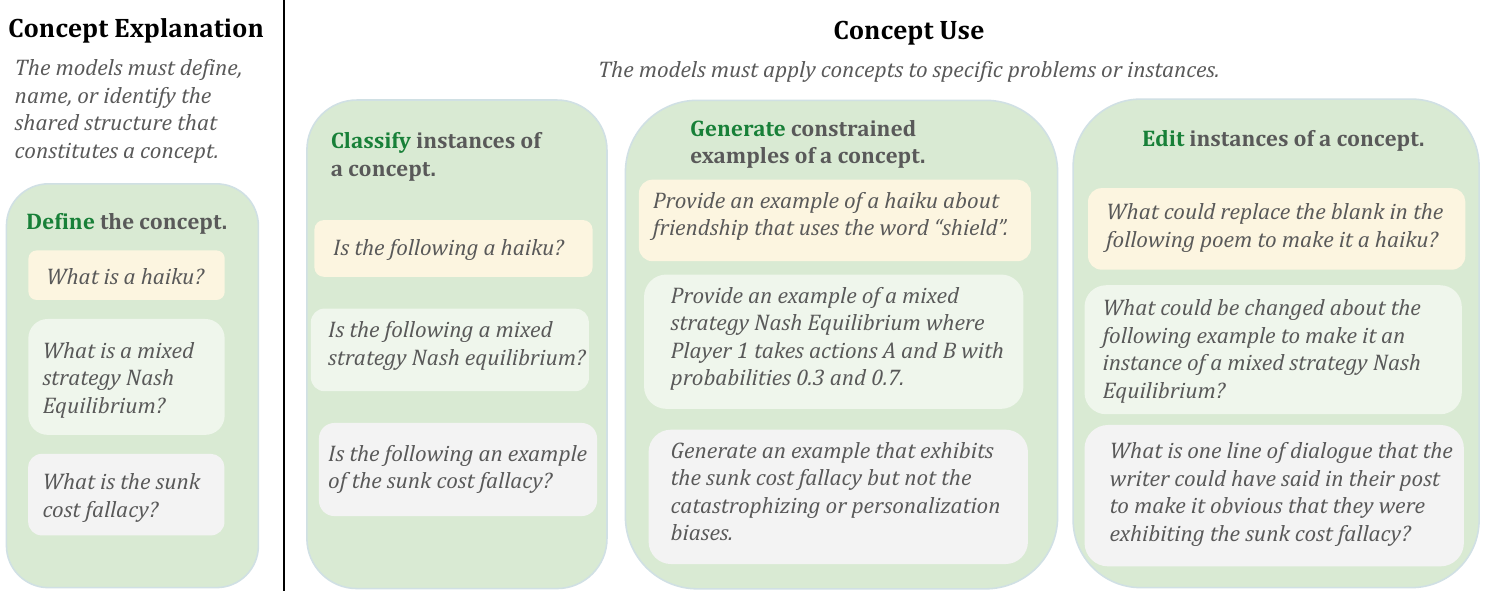}
    \caption{\emph{Our framework for evaluating concept understanding, including examples.}  We evaluate concept explanation by asking the model to define the concept. We measure concept use by giving the model classification, constrained generation, and edit tasks.}
    \label{fig:framework} \end{figure*}

Figure \ref{fig:framework} provides a schematic representation of our experimental framework, with examples from each of our three domains. The yellow examples correspond to the concept of a haiku, in the literary techniques domain. The green examples correspond to the concept of a mixed strategy Nash Equilibrium, in the game theory domain. Finally, the gray examples correspond to the concept of a sunk cost fallacy, in the psychological biases domain.

\newpage
\section{Concept Choices}
\label{appendix:concept_choices}
Within each domain, we selected a diverse set of concepts for our main analysis, totaling 32 concepts: 12 from literary techniques, 9 from game theory, and 11 from psychological biases. \textbf{Table \ref{tab:concepts}} lists all concepts with abbreviated definitions.

\begin{table}[H]
\rowcolors{1}{marinalight}{marinadark!20}
\footnotesize
\renewcommand{\arraystretch}{1.5} \begin{tabular}{|p{2cm}|p{6cm}|p{8cm}|}
\hline
\cellcolor{domains}\textbf{Domain} & \cellcolor{domains}\textbf{Concept} & \cellcolor{domains}\textbf{Definition} \\ \hline
& Haiku & A 3-line poem with a 5-7-5 syllable structure \\ \cline{2-3}
& Shakespearean Sonnet & A 14-line poem with ABABCDCDEFEFGG rhyme scheme, written in iambic pentameter \\ \cline{2-3}
& Analogy & Comparison highlighting similarities between two things \\ \cline{2-3}
& Paradox & A seemingly contradictory but true statement \\ \cline{2-3}
\textbf{Literary}& Anacoluthon & Sudden break in grammatical structure for effect \\ \cline{2-3}
\textbf{Techniques} & Asyndeton & Omission of conjunctions for a concise effect \\ \cline{2-3}
& Hyperbaton & Reordering words for emphasis \\ \cline{2-3}
& Synesis & Agreement of meaning over grammar \\ \cline{2-3}
& Accismus & Feigned refusal of something desired \\ \cline{2-3}
& Slant Rhyme & Near rhyme with similar but not identical sounds \\ \cline{2-3}
& Enthymeme & An argument with an implied premise \\ \cline{2-3}
& Anapest & Metrical foot with two unstressed and one stressed syllable \\ \hline

& Strict Dominance & A strategy always better than others, regardless of opponents \\ \cline{2-3}
& Iterated Dominance & Successive elimination of dominated strategies \\ \cline{2-3}
& Weak Dominance & A strategy at least as good as others and better in some cases \\ \cline{2-3}
\textbf{Game} & Pure Nash Equilibrium & Players choose specific strategies with no incentive to deviate \\ \cline{2-3}
\textbf{Theory} & Mixed Strategy Nash Equilibrium & Players randomize strategies with no incentive to deviate \\ \cline{2-3}
& Pareto Optimality & No player can be better off without making another worse off \\ \cline{2-3}
& Best Response & Choosing your payoff-maximizing response to others' actions \\ \cline{2-3}
& Zero-Sum Game & One player’s gain equals another’s loss \\ \cline{2-3}
& Symmetric Game & Payoffs depend on strategies, not players \\ \hline
& Fundamental Attribution Error & Explaining others' behavior by personality, not situation \\ \cline{2-3}
& Black and White Thinking & Viewing situations in extreme, all-or-nothing terms \\ \cline{2-3}
& Sunk Cost Fallacy & Irrationally continuing an endeavor due to past investments \\ \cline{2-3}
& IKEA Effect & Valuing something more because you assembled it yourself \\ \cline{2-3}
\textbf{Psychological} & Pseudocertainty Effect & Treating conditional outcomes as certain \\ \cline{2-3}
\textbf{Biases} & Endowment Effect & Valuing something more just because you own it \\ \cline{2-3}
& Naive Cynicism & Assuming others' motives are more self-interested than your own \\ \cline{2-3}
& Normalcy Bias & Underestimating risk by assuming things will continue as usual \\ \cline{2-3}
& Spotlight Effect & Overestimating how much others notice you \\ \cline{2-3}
& Illusory Superiority & Overestimating your abilities compared to others \\ \cline{2-3}
& Catastrophizing & Assuming the worst possible outcome will occur \\ 
\hline
\end{tabular}
\caption{Concepts and their definitions grouped by domain.}
\label{tab:concepts}
\end{table}

\newpage
\section{Specifications of the Classification Task}
\label{sec:app_specifications_classification}
After assembling the dataset of true and false instances of each concept, we automated the evaluation of model classifications by directly comparing the model’s outputs to our labeled ground truth. Across all of our models and domains, we collected a total of \textbf{2,030 annotations}. These annotations were collected as follows:
\begin{enumerate}
    \item \textbf{Literary techniques.} We evaluate $7$ models across $12$ concepts. For each concept, we consider $10$ examples—$5$ true instances and $5$ false instances. This results in a total of \textbf{840 annotations}.
    \item \textbf{Game theory.} We evaluate $7$ models across $9$ concepts. For each concept, we consider $10$ examples—$5$ true instances and $5$ false instances. This results in a total of \textbf{630 annotations}.
    \item \textbf{Psychological biases.} We gathered $40$ examples from posts on the subreddit \texttt{r/AmIOverreacting}, each annotated for the presence of $2$ (from a list of $11$) psychological biases. Labels were determined by majority vote among trained psychologists recruited via Upwork. Each psychologist evaluated $10$ posts, providing labels for $2$ biases per post. For more details on the survey design, see Appendix Section \ref{sec:app_expert_psych_survey}. Every (post, bias) pair was labeled by $3$ psychologists, and we retained the majority label. Subsequently, we asked each of $7$ models to classify each post for both biases. This results in a total of \textbf{560 annotations}.
\end{enumerate}
For our analysis, we exclude responses missing the specified format (e.g., missing strings like ``ANSWER: \_'').

\newpage
\section{Classification Task: Survey for Expert Annotation of Psychological Biases}
\label{sec:app_expert_psych_survey}
\begin{figure*}[h] \centering
    \includegraphics[width=0.5\textwidth]{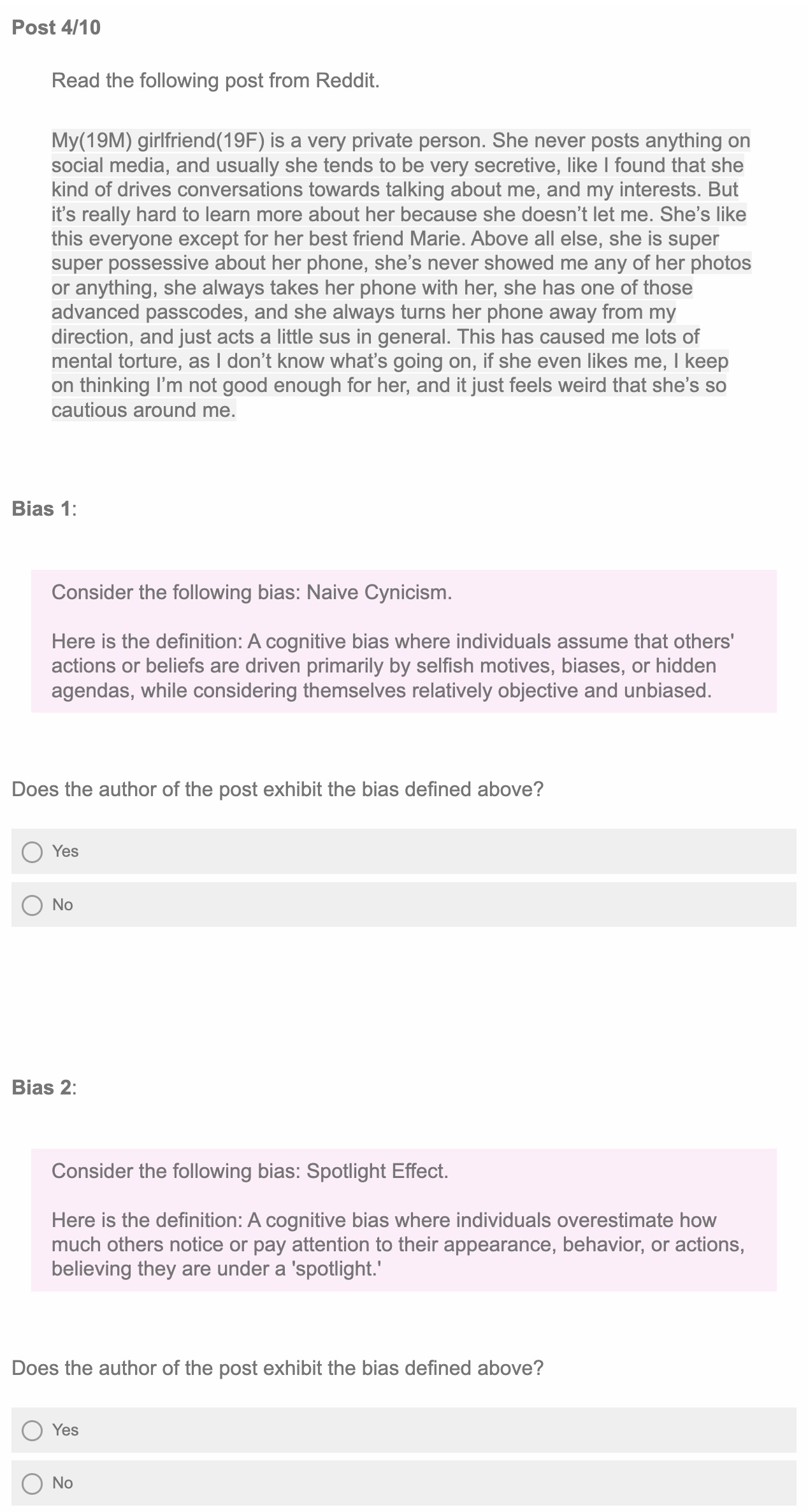}
    \caption{Example screen from the expert annotation survey used to label psychological biases.}
    \label{fig:survey_expert_labels_psych} \end{figure*}

To establish reliable ground-truth labels for psychological biases, we conducted an expert annotation survey with psychologists recruited via Upwork. Experts were shown posts from the subreddit \texttt{r/AmIOverreacting} and asked to determine whether each of two psychological biases was present in each post. Full specifications of the annotation collection process can be found in Appendix Section \ref{sec:app_specifications_classification}. Figure \ref{fig:survey_expert_labels_psych} shows an example screen from the expert annotation survey.

\newpage
\section{Specifications of the Constrained Generation Task}
\label{sec:app_specifications_generation}
\subsection{Literary Techniques}
In the domain of \emph{literary techniques}, we required model outputs to adhere to three types of constraints. The first constraint is \emph{semantic}: the example must involve a theme such as “gratitude” or “friendship.” The second constraint is \emph{form}: the example must comply with a linguistic constraint like using dialogue or starting with a monosyllabic word. The last constraint is \emph{lexical}: the example must incorporate a specific word, such as “gravity” or “wake.” Each prompt was constructed by randomly selecting one constraint from each category. For example, one prompt might be: “\emph{Write an example involving friendship. The example must use dialogue and include the word ‘gravity’}”. 

To generate these constraints, we prompted GPT-4o to produce a list of $20$ constraints for each of the three categories, totaling $60$ unique constraints. The full list of constraints is shown in Table \ref{tab:constraints_lit}. We consider $7$ models and $12$ concepts. In our analysis, the paper authors annotate one responses for each (concept, model) pair, resulting in \textbf{84 annotations}.

\begin{table}[h]
    \centering
    \begin{tabular}{lll}
        \toprule
        \textbf{Semantic} & \textbf{Form} & \textbf{Lexical} \\
        \midrule
        Nature & Start with a monosyllabic word. & Father \\
        Childhood memories & Start with a word that is not monosyllabic. & Red \\
        A significant event & Start with a word that begins with the letter "M". & Order \\
        Love & Start with a word that begins with the letter "I". & Press \\
        Happiness & Use alliteration at least once. & Jump \\
        Friendship & End with a one-vowel word. & Mother \\
        Overcoming adversity & Use dialogue. & Run \\
        Gratitude & Do not use dialogue. & Gravity \\
        Change & Do not use any adjectives. & Shine \\
        Hope & Do not use any adverbs. & Wonder \\
        Moments in time & Make every sentence start with a verb. & Just \\
        A meaningful object & Use the first-person perspective. & Justice \\
        A meaningful person & Use the second-person perspective. & King \\
        The earth & Use the third-person perspective. & Earth \\
        A hobby & Include a sentence that contains a question. & Sunshine \\
        Your dreams & Include a repeated phrase. & Light \\
        Loss & Have at least two sentences or lines that start with the same word. & Wake \\
        Your fears & Use a comma in the first sentence or line. & Friend \\
        Feeling lost & Do not use any pronouns. & Nectar \\
        Anger & Make the result in the past tense. & Love \\
        \bottomrule
    \end{tabular}
    \caption{\emph{A full list of constraints used for the literary techniques domain in the constrained generation task.} We generated model prompts by \textbf{independently sampling one random constraint from each column}.}
    \label{tab:constraints_lit}
\end{table}

\newpage
\subsection{Game Theory}
In the domain of \emph{game theory}, the constraints differed depending on the concept being tested. For instance, for the concept of ``strict dominance,'' a constraint could specify whether one or both players must have a strictly dominant strategy in the example. Similarly, for the concept of “Pareto optimality,” a constraint might require that all numbers in the payoff matrix be unique. We considered $7$ models and $9$ concepts. We generated one response for each (concept, model) pair, resulting in \textbf{63 annotations}. All annotations were generated by custom, automated evaluators.

\begin{table}[h]
    \centering
    \begin{tabular}{p{0.3\linewidth} p{0.6\linewidth}}
        \toprule
        \textbf{Concept Group} & \textbf{Constraint Prompt} \\
        \midrule
        \multirow{3}{*}{} 
        Strict Dominance & There is a $x$ strategy for (one/both) players. \\
        Weak Dominance & In the (row/column) that is $x$ for Player (1/2) receives a payoff of $a$ if the other player takes the corresponding actions. \\
        Pure Strategy Nash Equilibrium & In the payoff matrix, all numbers must be unique. \\
        \midrule
        Iterated Dominance & Start with an $a \times b$ matrix and end with a $c \times d$ matrix. \\
        \midrule
        \multirow{4}{*}{} 
        Zero-Sum Game& If Player (1/2) takes the $f$ action, they should receive a payoff of $(a, b, c)$ if the other player takes the corresponding actions. \\
        & In the payoff matrix, all numbers must be unique. \\
        \midrule
        Mixed Strategy Nash Equilibrium 
        & In the mixed strategy Nash equilibrium, Player (1/2) should take Action 1 with a probability of $p$ and Action 2 with a probability of $1-p$. \\
        \midrule
        Pareto Optimality 
        & You must generate a 3x3 matrix where there are exactly $e$ distinct Pareto optimal solutions. \\
        & In the payoff matrix, all numbers must be unique.\\
        \midrule
        Best Response
        & In the game, imagine Player 1 chooses the first action with a probability of $p_1$, the second action with a probability of $p_2$, and the third action with a probability of $p_3$. Player 2's best response to this action should be the $f$ action.  \\
        Symmetric Game & In the payoff matrix (combined for both players), all numbers must be unique, regardless of which Player the payoff is associated with. \\
        \bottomrule
    \end{tabular}
    \caption{\emph{A full list of constraints used for the game theory domain in the constrained generation task.} We generated model prompts differently for each concept group, applying all respective constraint prompts. When generating prompts, we randomly chose between items in parentheses, replaced items $x$ with versions of the concept name, probabilities $p$ with single-point decimal values between $0$ and $1$, values $a$ through $d$ with integers $\in [1, 9]$, value $e$ with an integer $\in [1, 5]$, and action $f \in \{\text{1st}, \text{2nd}, \text{3rd}\}$. Note that if applicable, the uniqueness constraints are present in each generation prompt with a probability of $\frac{1}{2}$.}
    \label{tab:game_theory_constraints}
\end{table}

\subsection{Psychological Biases}
In the \emph{psychological biases} domain, models were tasked with generating examples illustrating two specified biases while explicitly avoiding a third, randomly selected bias. Six expert psychologists, recruited via Upwork, annotated the resulting model responses using a Qualtrics survey. Each expert evaluated responses from all $7$ models across $2$ concepts per model. For more details on the Qualtrics survey, see Appendix Section \ref{sec_app:psych_generate_survey}. We obtained one response for each combination of $11$ concepts and $7$ models, resulting in \textbf{77 annotations}.

\newpage
\section{Generation Task: Survey for Expert Annotation of Psychological Biases}
\label{sec_app:psych_generate_survey}

\begin{figure*}[h] \centering
    \includegraphics[width=0.5\textwidth]{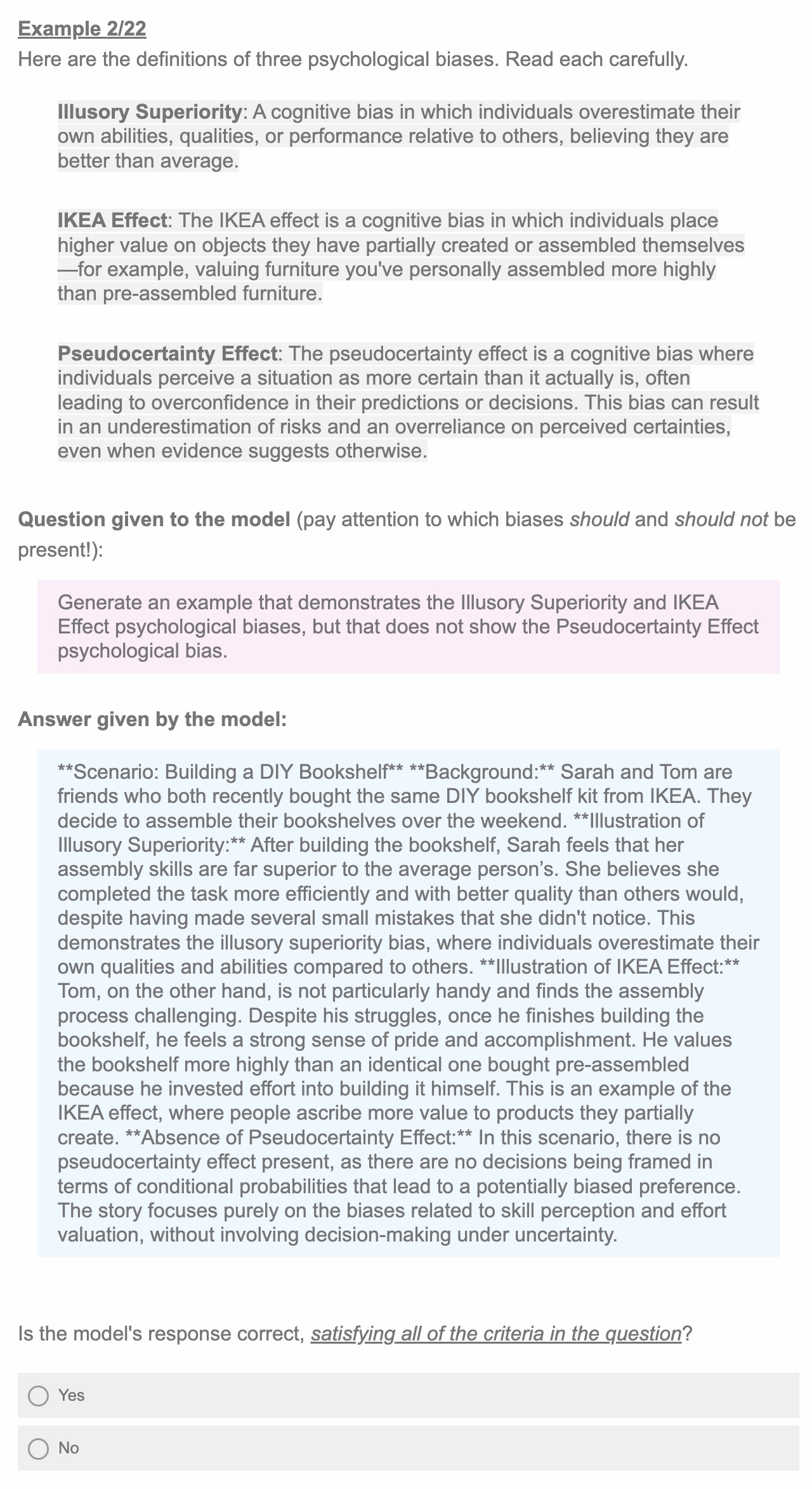}
    \caption{Example screen from the survey about evaluating model performance on our constrained generation task.}
    \label{fig:survey_screenshot_cgt} \end{figure*}

To establish reliable ground-truth labels for psychological biases, we conducted an expert annotation survey with professional psychologists recruited via Upwork. Experts were shown inferences where models were tasked with generating examples illustrating two specified biases while explicitly avoiding a third, randomly selected bias. Full specifications of the annotation collection process can be found in Appendix Section \ref{sec:app_specifications_generation}. Figure \ref{fig:survey_screenshot_cgt} shows an example screen from the expert annotation survey.

\newpage
\section{Specifications of the Editing Task}
\label{sec:app_specifications_editing}
In the domain of \emph{literary techniques}, we created ``masked'' versions of each example by removing key words or phrases necessary for satisfying each concept. For instance, a masked example of an ``analogy'' might omit one of the compared terms. Each model was tasked with editing these masked examples to restore them as either positive or negative instances. Model edits were evaluated by the paper authors, as certain concepts required specialized knowledge and adherence to specific constraints. For example, a valid Shakespearean sonnet must consist of $14$ lines in iambic pentameter following an ABABCDCDEFEFGG rhyme scheme. We evaluated edits by $7$ models across 12 concepts, resulting in \textbf{84 annotations}.

In the domain of \emph{game theory}, we presented each model with examples from our dataset and asked them to suggest modifications that would convert positive examples into negative ones or vice versa. For instance, we might provide a scenario that does not qualify as a mixed strategy Nash equilibrium and prompt the model to identify changes that would make it qualify. Model responses were automatically labeled as valid or invalid instances of the respective concepts using fully-automated, custom evaluation functions. We automatically validated model responses $10$ times for each (model, concept). Across $7$ models and $9$ concepts, this yielded a total of \textbf{630 annotations}.

In the domain of \emph{psychological biases}, models were prompted to generate a single line of dialogue that, when added to a Reddit post from our dataset, would either introduce a specific bias if it was originally absent or remove it if originally present. Ground-truth labels were determined by majority expert annotation as described in Appendix Sections \ref{sec:app_specifications_classification} and \ref{sec:app_expert_psych_survey}. For instance, if a post clearly exhibited catastrophizing, models were asked to provide a line negating this bias. Six expert psychologists, recruited via Upwork, annotated the resulting model responses using a Qualtrics survey (see Appendix Section \ref{sec:app_psych_survey_edit}). Each expert evaluated responses from all $7$ models across $2$ concepts per model. All $11$ psychological biases were evaluated across all models, resulting in \textbf{77 annotations}.

\newpage
\section{Editing Task: Survey for Expert Annotation of Psychological Biases}
\label{sec:app_psych_survey_edit}
\begin{figure*}[h] \centering
    \includegraphics[width=0.5\textwidth]{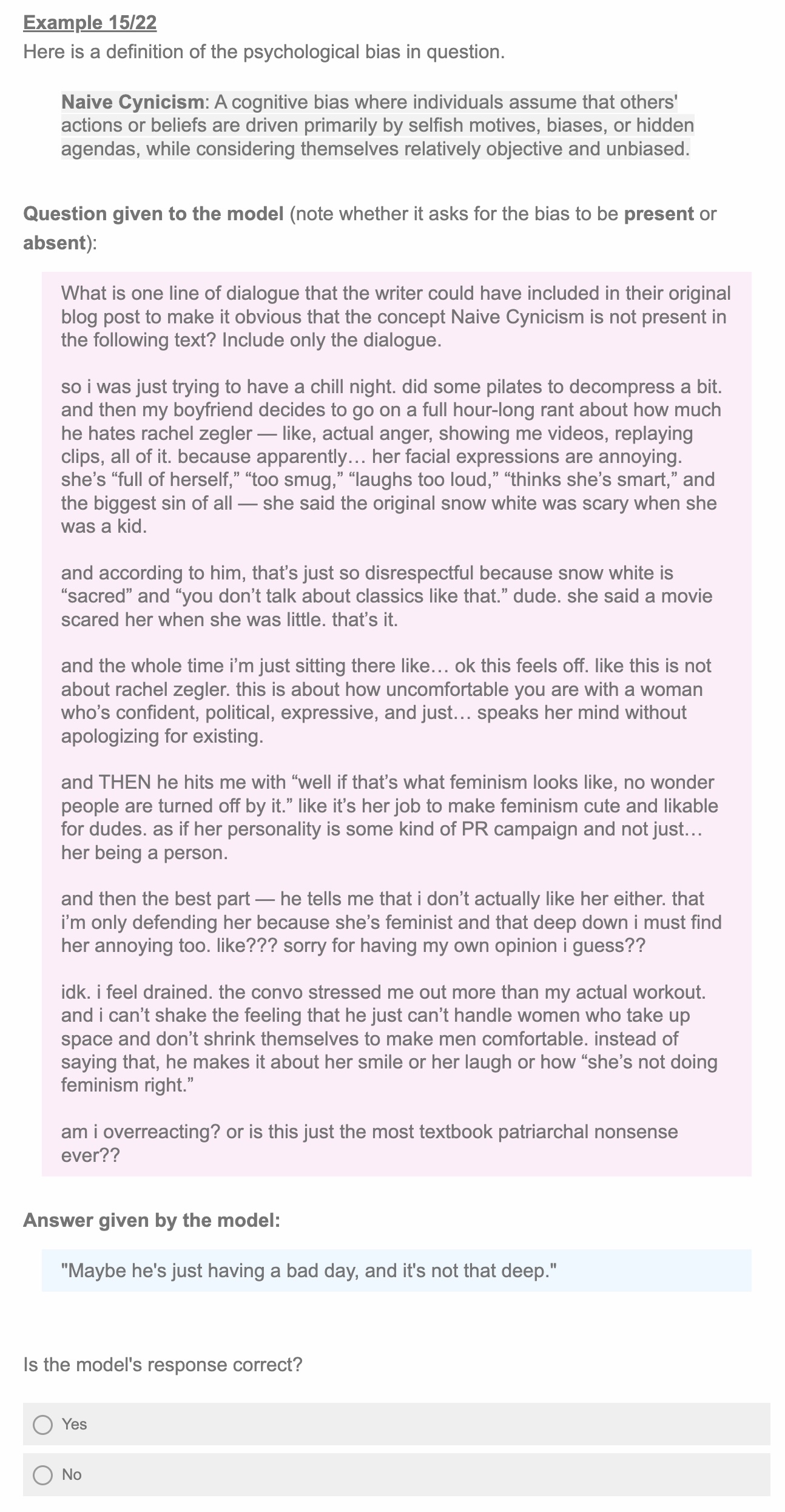}
    \caption{Example screen from the survey about evaluating model performance on our edit task.}
    \label{fig:survey_screenshot_edit} \end{figure*}

To establish reliable ground-truth labels for psychological biases, we conducted an expert annotation survey with professional psychologists recruited via Upwork. Experts were shown inferences where models were tasked with generating a single line of dialogue that, when added to a Reddit post from our dataset, would either introduce a specific bias if it was originally absent or remove it if originally present. Full specifications of the annotation collection process can be found in Appendix Section \ref{sec:app_specifications_editing}. Figure \ref{fig:survey_screenshot_edit} shows an example screen from the expert annotation survey.

\newpage
\section{Full Results from Benchmark Dataset Table}
\label{sec:app_full_benchmark_results}

Table \ref{tab:main_results_2} lists the results of our benchmark dataset, subdivided by domain. Conditioned on a correct definition for each (model, concept) pair, we evaluate accuracy levels in classification, generation, and editing tasks.

\begin{table*}[h!] \centering
\begin{tabular}{llccc}
\toprule
\textbf{Domain} & \textbf{Model} & \textbf{Classify} & \textbf{Generate} & \textbf{Edit} \\ \midrule
Literary Techniques
& Llama-3.3        & 0.64 (0.09) & 0.64 (0.15) & 0.27 (0.13) \\
& Claude-3.5       & 0.47 (0.08) & 0.42 (0.14) & 0.33 (0.14) \\
& GPT-4o           & 0.49 (0.08) & 0.25 (0.13) & 0.25 (0.13) \\
& Gemini-2.0       & 0.62 (0.09) & 0.17 (0.11) & 0.25 (0.13) \\
& DeepSeek-V3      & 0.56 (0.09) & 0.25 (0.13) & 0.25 (0.13) \\
& DeepSeek-R1      & 0.40 (0.08) & 0.42 (0.14) & 0.58 (0.14) \\
& Qwen2-VL         & 0.65 (0.11) & 0.67 (0.16) & 0.44 (0.17) \\ \midrule
Game Theory
& Llama-3.3        & 0.47 (0.09) & 0.56 (0.17) & 0.39 (0.05) \\
& Claude-3.5       & 0.42 (0.09) & 0.22 (0.14) & 0.31 (0.05) \\
& GPT-4o           & 0.49 (0.09) & 0.78 (0.14) & 0.36 (0.05) \\
& Gemini-2.0       & 0.36 (0.08) & 1.00 (0.00) & 0.50 (0.05) \\
& DeepSeek-V3      & 0.47 (0.09) & 0.78 (0.14) & 0.40 (0.05) \\
& DeepSeek-R1      & 0.18 (0.07) & 0.63 (0.17) & 0.50 (0.06) \\
& Qwen2-VL         & 0.53 (0.09) & 1.00 (0.00) & 0.51 (0.05) \\ \midrule
Psychological Biases
& Llama-3.3        & 0.62 (0.11) & 0.10 (0.10) & 0.20 (0.13) \\
& Claude-3.5       & 0.62 (0.11) & 0.00 (0.00) & 0.00 (0.00) \\ 
& GPT-4o           & 0.62 (0.11) & 0.18 (0.12) & 0.36 (0.15) \\
& Gemini-2.0       & 0.65 (0.11) & 0.18 (0.12) & 0.09 (0.09) \\
& DeepSeek-V3      & 0.70 (0.11) & 0.18 (0.12) & 0.18 (0.12) \\
& DeepSeek-R1      & 0.82 (0.12) & 0.18 (0.12) & 0.55 (0.15) \\
& Qwen2-VL         & 0.82 (0.12) & 0.27 (0.13) & 0.64 (0.15) \\ \midrule
\textbf{Overall}& \textbf{Overall} & \textbf{0.55 (0.02)} & \textbf{0.40 (0.03)} & \textbf{0.40 (0.02)} \\
\bottomrule
\end{tabular}
\caption{\emph{Potemkin rate on ``use'' tasks in our three selected domains}, conditioned on correctly defining each concept. Since random‐chance accuracy on classification is $0.5$ (implying baseline potemkin rate of $0.5$), we multiply the classification values by $2$ to rescale them to the same range as the other tasks. Standard errors are in parentheses.}
\label{tab:main_results_2}
\end{table*}

\newpage
\section{Qualitative Examples of Potemkins}
\label{sec:app_qualitative_potemkins}
In this section, we present qualitative examples of potemkins from various models, specifically selected to illustrate mistakes humans typically would not make. The examples contain abbreviated versions of model responses.

\begin{figure}[h!]
    \centering
    \begin{tabular}{cc}
        \includegraphics[width=0.5\textwidth]{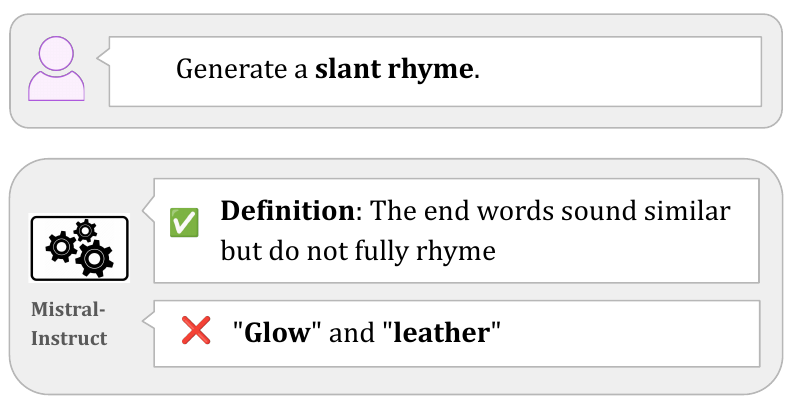} &
        \includegraphics[width=0.5\textwidth]{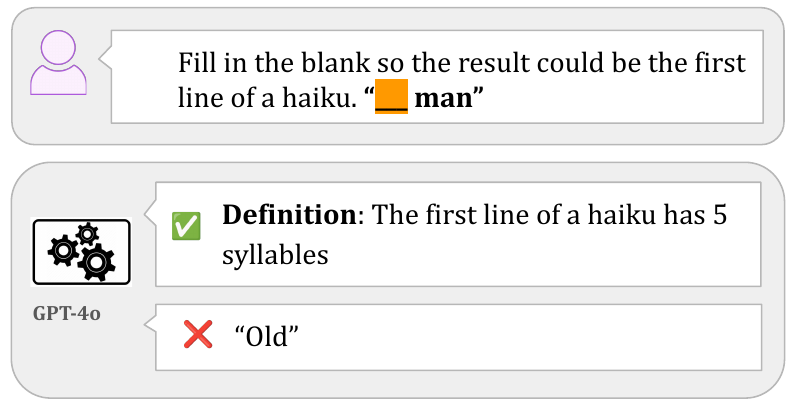} \\
        (a) & (b) \\[8pt]
        \includegraphics[width=0.5\textwidth]{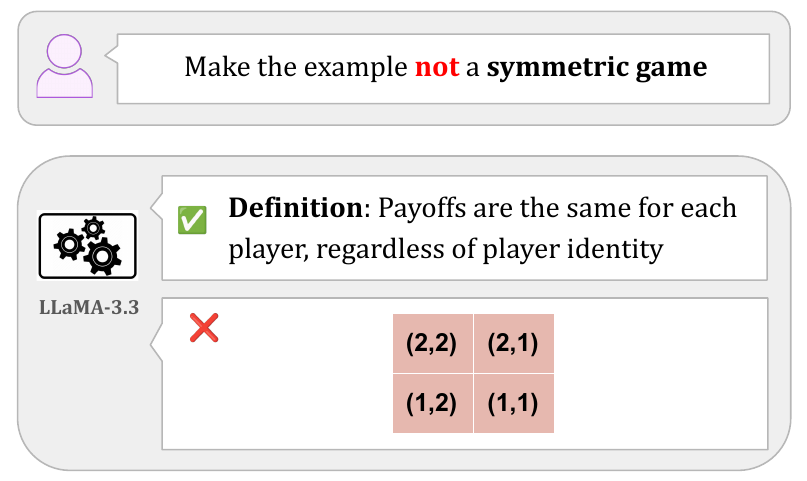} &
        \includegraphics[width=0.5\textwidth]{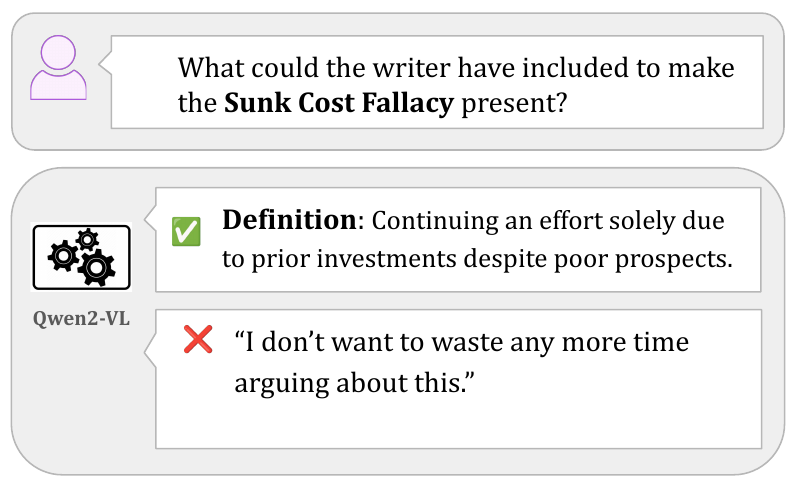} \\
        (c) & (d)
    \end{tabular}
    \caption{Selected qualitative examples of potemkins across models and domains.}
    \label{fig:qualitative_potemkins}
\end{figure}

\newpage
\section{Incoherence Data Collection Details}
\label{sec:app_incoherence_data_collection}
For our incoherence analysis, we evaluate $9$ models using the concepts from our three selected domains. Specifically, we prompt each model to generate $5$ true and $5$ false instances for each of our $32$ concepts. This results in a total of $2,880$ labeled examples: $1,080$ from the domain of literary techniques, $810$ from game theory, and $990$ from psychological biases.

Our analysis spans the following $9$ models: Llama-3.3 (70B), GPT-4o, GPT-o1-mini, GPT-o3-mini, Gemini-2.0 (Flash), Claude-3.5 (Sonnet), DeepSeek-V3, DeepSeek-R1, and Qwen2-VL(72B).

\section{Incoherence Scores by Domain}
\label{sec:app_incoherence_by_domain}

Table \ref{tab:coherence_table_by_domain} shows the incoherence scores of the models, broken down by domain.

\begin{table*}[h]
\centering
\begin{tabular}{lcccc}
\toprule
\textbf{Model} & \textbf{Literary Techniques} & \textbf{Game Theory} & \textbf{Psychological Biases} & \textbf{Overall} \\
\midrule
Llama-3.3     & 0.32 (0.06) & 0.24 (0.06) & 0.00 (0.00) & \textbf{0.19 (0.03)} \\
Claude-3.5    & 0.44 (0.08) & 1.04 (0.10) & 0.44 (0.08) & \textbf{0.61 (0.05)} \\
GPT-4o        & 0.70 (0.08) & 0.88 (0.10) & 0.38 (0.08) & \textbf{0.64 (0.05)} \\
GPT-o1-mini   & 0.28 (0.06) & 0.09 (0.04) & 0.07 (0.04) & \textbf{0.16 (0.03)} \\
GPT-o3-mini   & 0.05 (0.03) & 0.02 (0.02) & 0.00 (0.00) & \textbf{0.03 (0.01)} \\
Gemini-2.0    & 0.12 (0.04) & 0.16 (0.06) & 0.02 (0.02) & \textbf{0.09 (0.02)} \\
DeepSeek-V3   & 0.16 (0.06) & 0.18 (0.06) & 0.04 (0.02) & \textbf{0.13 (0.03)} \\
DeepSeek-R1   & 0.04 (0.02) & 0.08 (0.04) & 0.00 (0.00) & \textbf{0.04 (0.02)} \\
Qwen2-VL      & 0.16 (0.06) & 0.06 (0.04) & 0.12 (0.04) & \textbf{0.13 (0.03)} \\
\midrule
\textbf{Overall} & \textbf{0.25 (0.02)} & \textbf{0.31 (0.03)} & \textbf{0.12 (0.02)} & \textbf{0.22 (0.01)} \\
\bottomrule
\end{tabular}
\caption{\emph{Incoherence scores across domains and models.} A score of $0$ indicates perfect performance and a score of $1$ indicates good-as-random performance. Standard errors are in parentheses. }
\label{tab:coherence_table_by_domain}
\end{table*}

\newpage
\section{Prompts for Automatic Evaluation Procedure}
\label{sec:app_auto_eval_prompts}

In this section, we will list the prompts given to models during the various stages of the automatic evaluation procedure.

The following prompt is used for models to \textbf{generate questions}.

\begin{lstlisting}*[caption={Prompt to generate questions},label={lst:gen_questions_prompt}]
The following is a question about the following concept: {concept}.

Here is the question: {question}.

Write {num_subquestions} other questions that test whether someone who understands the concepts the question is testing truly understands them.
\end{lstlisting}

The following prompts are used for models to \textbf{generate correct and incorrect answers}, respectively.

\begin{lstlisting}*[caption={Prompt to generate correct answers},label={lst:gen_answers_prompt_correct}]
You are an expert tutor. You may think step-by-step to reach the solution, but you MUST finish with an answer that starts exactly with `{FINAL_TAG}` followed by your single best answer. If the question asks you to explain your answer, you should also include the explanation after `{FINAL_TAG}`.


Question:{question}
\end{lstlisting}

\begin{lstlisting}*[caption={Prompt to generate incorrect answers},label={lst:gen_answers_prompt_incorrect}]
Modify the following answer to introduce a subtle error. The error should be subtle but one such that a human who knows the concept would know the answer is incorrect. If the answer is already incorrect, you can leave it the same. You can reason all you'd like, but end the response with `{FINAL_TAG}` followed by the full modified answer.

Question: {question}
Answer: {initial_answer}
\end{lstlisting}

Finally, the following prompt is used for models to \textbf{judge answers}.
\begin{lstlisting}*[caption={Prompt to judge answers},label={lst:gen_judge_answers}]
You are an expert tutor. You will be given a question and a possible answer to the question. Your job is to determine if the answer is correct or incorrect. You should only grade it correct if the answer (including the reasoning) is completely correct. You can reason all you'd like, but end the response with `{FINAL_TAG}` followed by either 'correct' or 'incorrect', and nothing should come after that.


Question: {question}
Answer: {model_answer}
\end{lstlisting}

\end{document}